\documentclass{llncs}
\usepackage{bm}
\usepackage[utf8]{inputenc}
\usepackage{url}
\usepackage{amssymb}
\usepackage{amsfonts}
\usepackage{graphicx}
\usepackage{url}
\usepackage{lstsemantic}
\usepackage{lstlangmizar}
\usepackage{booktabs}
\usepackage{footnoterange}

\usepackage{float}
\floatstyle{boxed}
\restylefloat{figure}

\usepackage{hyperref}

\title{ProofWatch:\\Watchlist Guidance for Large Theories in E}

\author{Zarathustra Goertzel\inst{1} \and Jan Jakub\r{u}v\inst{1} \and Stephan Schulz\inst{2}
  \and Josef Urban\inst{1}\thanks{Supported by the \textit{AI4REASON} ERC Consolidator grant number 649043, and by the Czech project AI\&Reasoning CZ.02.1.01/0.0/0.0/15\_003/0000466 and the European Regional Development Fund.}}
\institute{
  Czech Technical University in Prague
\and
  DHBW Stuttgart
}

\newcommand*{\decayrel}{\mathop{\mathit{relevance}_1}}
\newcommand*{\peparel}{\mathop{\mathit{relevance}_2}}
\newcommand*{\progress}{\mathop{\mathit{progress}}}
\newcommand*{\parents}{\mathop{\mathit{parents}}}
\newcommand*{\stdrel}{\mathop{\mathit{relevance}_0}}

\begin{document}

\maketitle{}

\begin{abstract}
  Watchlist (also hint list) is a mechanism that allows related proofs
  to guide a proof search for a new conjecture. This mechanism has
  been used with the Otter and Prover9 theorem provers, both for
  interactive formalizations and for human-assisted proving of open
  conjectures in small theories.  In this work we
  explore %
  the use of watchlists in large theories coming from first-order
  translations of large ITP libraries, aiming at improving
  hammer-style automation by smarter internal guidance of the ATP
  systems. In particular, we
  (i) design watchlist-based clause evaluation heuristics inside the E
  ATP system, and (ii) develop new proof guiding algorithms that load
  many previous proofs inside the ATP and focus the proof search using
  a dynamically updated notion of proof matching.  The methods are
  evaluated on a large set of problems coming from the Mizar library,
  showing significant improvement of E's standard portfolio of
  strategies, and also of the previous best set of strategies invented
  for Mizar by evolutionary methods.
\end{abstract}

\section{Introduction: Hammers, Learning and Watchlists}
\label{Intro}
\emph{Hammer}-style automation tools connecting
interactive theorem provers (ITPs) with automated theorem provers
(ATPs) have recently led to a significant speed\-up for formalization
tasks~\cite{hammers4qed}.
An important component of such tools is \emph{premise
  selection}~\cite{abs-1108-3446}: choosing a small number of the most
relevant facts that are given to the
ATPs. %
Premise selection methods based on machine learning from many proofs
available in the ITP libraries typically outperform manually
specified %
heuristics~\cite{abs-1108-3446,holyhammer,KaliszykU13b,FarberK15,BlanchetteGKKU16,IrvingSAECU16}.
Given the performance of such %
\emph{ATP-external guidance} methods, learning-based \emph{internal
  proof search guidance} methods have started to be explored, both for
ATPs~\cite{UrbanVS11,KaliszykU15,JakubuvU17a,LoosISK17,FarberKU17} and
also in the context of tactical ITPs~\cite{GauthierKU17,GransdenWR15}.

In this work we develop learning-based internal proof guidance methods
for the E~\cite{Schulz13} ATP system and evaluate them on the large
Mizar Mathematical Library~\cite{mizar-in-a-nutshell}.  The methods
are based on the \emph{watchlist} (also \emph{hint list}) technique
developed by Veroff~\cite{Veroff:JAR-1996}, focusing proof search
towards lemmas (\emph{hints}) that were useful in related
proofs. Watchlists have proved essential in the AIM
project~\cite{KinyonVV13} done with Prover9~\cite{McCune:WWW-2008} for
obtaining very long and advanced proofs of open conjectures.  Problems
in large ITP libraries however differ from one another much more than
the AIM problems, making it more likely for unrelated watchlist lemmas
to mislead the proof search. Also, Prover9 lacks a number of
large-theory mechanisms and strategies developed recently for
E~\cite{KaliszykSUV15,JakubuvU16,JakubuvU17a}.

Therefore, we first %
design watchlist-based clause evaluation heuristics for E that can be
combined with other E strategies. Second, we complement the internal
watchlist guidance by using external statistical machine learning to
pre-select smaller numbers of watchlist clauses %
relevant for the current problem. Finally, we use the watchlist
mechanism to develop new proof guiding algorithms that load
many previous proofs inside the ATP and focus the search using a
  \emph{dynamically} updated heuristic representation of \emph{proof search state} based on matching the previous proofs.

  The rest of the paper is structured as follows. Section~\ref{Basics} briefly summarizes the work of saturation-style
ATPs such as E. Section~\ref{State}
discusses heuristic representation of search
  state and its importance for learning-based proof guidance. We
  propose an abstract vectorial representation expressing similarity to
  other proofs as a suitable evolving characterization of saturation
  proof searches. We also propose a concrete implementation based on
  \emph{proof completion ratios} tracked by the watchlist mechanism.
  Section~\ref{WatchE}
describes the standard (\emph{static}) watchlist
  mechanism implemented in E and Section~\ref{Dynamic} introduces the
  new \emph{dynamic} watchlist mechanisms and its use for guiding the
  proof search. Section~\ref{ExpStatic} evaluates the static
  and dynamic watchlist guidance combined with learning-based
  pre-selection on the Mizar library.
 Section~\ref{Examples} shows
  several examples of nontrivial proofs obtained by the new methods,
  and Section~\ref{Related} discusses related work and possible
  extensions.

\section{Proof Search in Saturating First-Order Provers}
\label{Basics}

The state of the art in first-order theorem proving is a saturating
prover based on a combination of resolution/paramodulation and
rewriting, usually implementing a variant of the superposition
calculus~\cite{BG94}. In this model, the \emph{proof state} is represented as
a set of first-order clauses (created from the axioms and the negated
conjecture), and the system systematically adds  logical consequences
to the state, trying to derive the empty clause and hence an explicit
contradiction.

All current saturating first-order provers are based on variants of
the \emph{given-clause algorithm}. In this algorithm, the proof state
is split into two subsets of clauses, the processed clauses $P$
(initially empty) and the unprocessed clauses $U$. On each iteration
of the algorithm, the prover picks one unprocessed clause $g$ (the
so-called \emph{given clause}), performs all inferences which are
possible with $g$ and all clauses in $P$ as premises, and then moves
$g$ into $P$. The newly generated consequences are added to $U$. This
maintains the core invariant that all inferences between clauses in
$P$ have been performed. Provers differ in how they integrate
simplification and redundancy into the system, but all enforce the
variant that $P$ is maximally simplified (by first simplifying $g$
with clauses in $P$, then back-simplifying $P$ with $g$) and that $P$
contains neither tautologies nor subsumed clauses.

The core choice point of the given-clause algorithm is the selection
of the next clause to process. If theoretical completeness is desired,
this has to be \emph{fair}, in the sense that no clause is delayed
forever. In practice, clauses are ranked using one or more heuristic
evaluation functions, and are picked in order of increasing evaluation
(i.e. small values are good). The most frequent heuristics are based
on symbol counting, i.e., the evaluation is the number of symbol occurrences
in the clause, possibly weighted for different symbols or symbols
types. Most provers also support interleaving a symbol-counting
heuristic with a first-in-first-out (FIFO) heuristic. E supports the
dynamic specification of an arbitrary number of differently
parameterized priority queues that are processed in weighted
round-robbin fashion via a small \emph{domain-specific language} for
heuristics.

Previous work~\cite{Schulz:KI-2001,SM:IJCAR-2016} has both shown that
the choice of given clauses is critical for the success rate of a
prover, but also that existing heuristics are still quite bad -
i.e. they select a large majority of clauses not useful for a given
proof. Positively formulated, there still is a huge potential for
improvement.

\section{Proof Search State in Learning Based Guidance}
\label{State}
A good representation of the current \emph{state} is crucial for
learning-based guidance. This is quite clear in theorem proving and
famously so in Go and Chess~\cite{SilverHMGSDSAPL16,abs-1712-01815}.
For example, in the TacticToe system~\cite{GauthierKU17} proofs are
composed from pre-pro\-gram\-med HOL4~\cite{SlindN08} tactics that are
chosen by statistical learning based on similarity of the evolving
\emph{goal state} to the goal states from related proofs. Similarly,
in the learning versions of leanCoP~\cite{OB03} --
(FE)MaLeCoP~\cite{UrbanVS11,KaliszykU15} -- the tableau extension
steps are guided by a trained learner using similarity of the evolving
tableau (the ATP \emph{proof search state}) to many other tableaux
from related proofs.

Such intuitive and compact notion of proof search state is however
hard to get when working with today's high-performance
saturation-style ATPs such as E~\cite{Schulz13} and
Vampire~\cite{Vampire}.  The above definition of saturation-style
proof state (Section~\ref{Basics}) as either one or two
(processed/unprocessed) large sets of clauses is very unfocused.
Existing learning-based guiding methods for
E~\cite{JakubuvU17a,LoosISK17} practically ignore
this. %
Instead, they use only the original conjecture and its features for
selecting the relevant given clauses throughout the whole proof search.

This is obviously unsatisfactory, both when compared to the evolving
search state in the case of tableau and tactical proving, and also
when compared to the way humans select the next steps when they search
for proofs. The proof search state in our mind is certainly an
evolving concept based on the search done so far, not a fixed set of
features extracted just from the conjecture.

\subsection{Proof Search State Representation for Guiding Saturation}

One of the motivations for the work presented here is to produce an
intuitive, compact and evolving heuristic representation of proof search state
in the context of learning-guided saturation proving. As usual, it
should be a vector of (real-valued) features that are
either manually designed or learned. In a high-level way, our proposed
representation is a \emph{vector expressing an abstract similarity of the
  search state to (possibly many) previous related proofs}. This can
be implemented in different ways, using both statistical and symbolic
methods and their combinations. An example and motivation comes again
from the work of Veroff, where a search is considered
promising when the given clauses frequently match hints. The gaps
between the hint matchings may correspond to the more brute-force
bridges between the different proof ideas expressed by the hints.

Our first practical implementation introduced in Section~\ref{Dynamic} is to load
upon the search initialization $N$ related proofs $P_i$, and for each $P_i$
keep track of the ratio of the clauses from $P_i$ that have already been subsumed during the search.
The subsumption checking is using E's watchlist
mechanism (Section~\ref{WatchE}).
The $N$-long vector $\bm{p}$ of such \emph{proof completion ratios} is our heuristic
representation of the proof search state, which is both compact and typically
evolving, making it suitable for both hard-coded and learned clause selection heuristics.

In this work we start with fast hard-coded watchlist-style heuristics for focusing inferences on clauses
that progress the more finished proofs (Section~\ref{Dynamic}). However training e.g. a statistical ENIGMA-style~\cite{JakubuvU17a}
clause evaluation model
by adding $\bm{p}$ to the currently used ENIGMA features is a straightforward extension.

\section{Static Watchlist Guidance and its Implementation in E}
\label{WatchE}

E originally implemented a watchlist mechanism as a means to force
direct, constructive proofs in first order logic. For this
application, the watchlist contains a number of goal clauses
(corresponding to the hypotheses to be proven), and all newly
generated and processed clauses are checked against the watchlist. If
one of the watchlist clauses is subsumed by a new clause, the former
is removed from the watchlist. The proof search is complete, once all
clauses from the watchlist have been removed. In contrast to the
normal proof by contradiction, this mechanism is not
complete. However, it is surprisingly effective in practice, and it
produces a proof by forward reasoning.

It was quickly noted that the basic mechanism of the watchlist can
also be used to implement a mechanism similar to the \emph{hints}
successfully used to guide Otter~\cite{MW:JAR-97} (and its successor
Prover9~\cite{McCune:WWW-2008}) in a semi-interactive
manner~\cite{Veroff:JAR-1996}. Hints in this sense are intermediate
results or lemmas expected to be useful in a proof. However, they are
not provided as part of the logical premises, but have to be derived
during the proof search. While the hints are specified when the prover
is started, they are only used to guide the proof search - if a clause
matches a hint, it is prioritized for processing. If all clauses
needed for a proof are provided as hints, in theory the prover can be
guided to prove a theorem without any search, i.e.\ it can
\emph{replay} a previous proof. A more general idea, explored in this
paper, is to fill the watchlist with a large number of clauses useful
in proofs of similar problems.

In E, the watchlist is loaded on start-up, and is stored in a feature
vector index~\cite{Schulz:FVI-2013} that allows for efficient retrieval
of subsumed (and subsuming) clauses. By default, watchlist clauses are
simplified in the same way as processed clauses, i.e. they are kept in
normal form with respect to clauses in $P$. This increases the chance
that a new clause (which is always simplified) can match a similar
watchlist clause. If used to control the proof search, subsumed
clauses can optionally remain on the watchlist.

We have extended E's domain-specific language for search heuristics
with two priority functions to access information about the
relationship of clauses to the watchlist - the function
\texttt{PreferWatchlist} gives higher rank to clauses that subsume at least one
watchlist clause, and the dual function
\texttt{DeferWatchlist} ranks them lower. Using the first, we have
also defined four built-in heuristics that preferably process
watchlist clauses. These include a pure watchlist heuristic, a simple
interleaved watch list function (picking 10 out of every eleven
clauses from the watchlist, the last using FIFO), and a modification
of a strong heuristic obtained from a genetic
algorithm~\cite{SchaferS15} that interleaves several different
evaluation schemes and was modified to prefer watchlist clauses in two
of its four sub-evaluation functions.

\section{Dynamic Watchlist Guidance}
\label{Dynamic}

In addition to the above mentioned \emph{static watchlist guidance},
we propose and experiment with an alternative: \emph{dynamic watchlist guidance}.
With dynamic watchlist guidance, several watchlists, as opposed to a single
watchlist, are loaded on start-up.  Separate watchlists are supposed to group
clauses which are more likely to appear together in a single proof.  The easiest
way to produce watchlists with this property is to collect previously proved
problems and use their proofs as watchlists. This is our current implementation, i.e., each watchlist
corresponds to a previous proof.
During a proof search, we maintain for each watchlist its \emph{completion status}, i.e.
the number of clauses that were already encountered.
The main idea behind our dynamic watchlist guidance is to prefer %
clauses which appear on watchlists that are closer to completion. Since watchlists now exactly correspond to previous
refutational proofs, completion of any watchlist implies that the current proof search is finished.

\subsection{Watchlist Proof Progress}

Let watchlists $W_1$,$\ldots$,$W_n$ be given for a proof search.
For each watchlist $W_i$ we keep a \emph{watchlist progress counter}, denoted
$\progress(W_i)$, which is initially set to $0$.
Whenever a clause $C$ is generated during the proof search, we have to check
whether $C$ subsumes some clause from some watchlist $W_i$.
When $C$ subsumes a clause from $W_i$ we increase $\progress(W_i)$ by
$1$.
The subsumed clause from $W_i$ is then marked as encountered, and it is not
considered in future watchlist subsumption checks.%
\footnote{
Alternatively, the subsumed watchlist clause $D\in W_i$ can be considered for
future subsumption checks but the watchlist progress counter $\progress(W_i)$
should not be increased when $D$ is subsumed again.  This is because we want the
progress counter to represent the number of \emph{different} clauses from $W_i$
encountered so far.}
Note that a single generated clause $C$ can subsume several clauses from one or
more watchlists, hence several progress counters might be increased multiple
times as a result of generating $C$.

\subsection{Standard Dynamic Watchlist Relevance}
\label{Dynamic1}

The easiest way to use progress counters to guide given clause selection is
to assign the \emph{(standard) dynamic watchlist relevance} to each generated clause
$C$, denoted $\stdrel(C)$, as follows.
Whenever $C$ is generated, we check it against all the watchlists for
subsumption and we update watchlist progress counters.
Any clause $C$ which does not subsume any watchlist clause is given
$\stdrel(C)=0$.
When $C$ subsumes some watchlist clause, its relevance is the maximum watchlist
completion ratio over all the matched watchlists.
Formally, let us write $C\sqsubseteq W_i$ when clause $C$ subsumes some clause from
watchlist $W_i$.
For a clause $C$ matching at least one watchlist, its relevance
is computed as follows.
\[
   \stdrel(C) =
   \max_{W\in
      \{W_i: C\sqsubseteq W_i\}
   }
   \Big(
      \frac{\progress(W)}{|W|}
   \Big)
\]
The assumption is that a watchlist $W$ that is matched more is more relevant to the current proof search.
In our current implementation, the relevance is computed at the time of
generation of $C$ and it is not updated afterwards.
As future work, we propose to also update the relevance of all generated but not yet
processed clauses from time to time in order to reflect updates of the watchlist
progress counters. Note that this is expensive, as the number of generated clauses is typically high. Suitable indexing
could be used to lower this cost or even to do the update immediately just for the affected clauses.

To use the watchlist relevance in E, we extend E's domain-specific language for
search heuristics with two priority functions \texttt{PreferWatchlistRelevant}
and \texttt{DeferWatchlistRelevant}.
The first priority function ranks higher the clauses with higher watchlist
relevance\footnote{
   Technically, E's priority function returns an integer priority, and
   clauses with smaller values are preferred.
   Hence we compute the priority as $1000*(1-\stdrel(C))$.
}, and the other function does the opposite.  These priority functions
can be used to build E's heuristics just like in the case of the static
watchlist guidance.
As a results, we can instruct E to process watchlist-relevant clauses in
advance.

\subsection{Inherited Dynamic Watchlist Relevance}
\label{Dynamic2}

The previous standard watchlist relevance prioritizes only clauses subsuming
watchlist clauses but it behaves indifferently with respect to other clauses.
In order to provide some guidance even for clauses which do not subsume any
watchlist clause, we can examine the watchlist relevance of the parents of each
generated clause, and prioritize clauses with watchlist-relevant parents.
Let $\parents(C)$ denote the set of previously processed clauses from which $C$
have been derived.
\emph{Inherited dynamic watchlist relevance}, denoted $\decayrel$, is a combination of
the standard dynamic relevance with the average of parents relevances multiplied by a
\emph{decay} factor $\delta<1$.
\[
   \decayrel(C) =
   \stdrel(C) + \delta*
   \mathop{\mathrm{avg}}\limits_{D\in\parents(C)}
   \big(
      \decayrel(D)
   \big)
\]
Clearly, the inherited relevance equals to the standard relevance for the
initial clauses with no parents.
The decay factor ($\delta$) determines the importance of parents watchlist
relevances.\footnote{In our experiments, we use $\delta=0.1$}
Note that the inherited relevances of $\parents(C)$ are already precomputed at
the time of generating $C$, hence no recursive computation is necessary.

With the above $\decayrel$ we compute the average of parents \emph{inherited}
relevances, hence the inherited watchlist relevance accumulates relevance of all
the ancestors.
As a result, $\decayrel(C)$ is greater than 0 if and only if $C$ has some
ancestor which subsumed a watchlist clause at some point.
This might have an undesirable effect that clauses unrelated to the watchlist
are completely ignored during the proof search.
In practice, however, it seems important to consider also watchlist-unrelated
clauses with some degree in order to prove new conjectures which do not appear
on the input watchlist.
Hence we introduce two \emph{threshold} parameters $\alpha$ and $\beta$ which
resets the relevance to 0 as follows.
Let $\mathit{length}(C)$ denote the length of clause $C$, counting occurrences of
symbols in $C$.
\[
   \peparel(C) = \left\{
   \begin{array}{ll}
      0 & \mbox{ iff }
         \decayrel(C) < \alpha
         \mbox{ and }
         \frac{\decayrel(C)}{\mathit{length}(C)} < \beta

   \\
      \decayrel(C) & \mbox{ otherwise}
   \end{array}\right.
\]
Parameter $\alpha$ is a threshold on the watchlist inherited relevance while $\beta$
combines the relevance with the clause length.\footnote{In our experiments, we
use $\alpha=0.03$ and $\beta=0.009$. These values have been found useful by a small grid search over a random sample of 500 problems.}
As a result, shorter watchlist-unrelated clauses are preferred to longer
(distantly) watchlist-related clauses.

\newpage
\section{Experiments with Watchlist Guidance}
\label{ExpStatic}

For our experiments we construct watchlists from
the %
proofs found by E on a benchmark of $57897$
Mizar40~\cite{KaliszykU13b} problems in the MPTP
dataset~\cite{Urban06}.\footnote{Precisely, we have used the small
  (\emph{bushy}, re-proving) versions, but without ATP
  minimization. They can be found at
  \url{http://grid01.ciirc.cvut.cz/~mptp/7.13.01_4.181.1147/MPTP2/problems_small_consist.tar.gz}} %
\footnote{Experimental results and code can be found at \url{https://github.com/ai4reason/eprover-data/tree/master/ITP-18}.}.
These initial proofs were found by an evolutionarily
optimized~\cite{JakubuvU17} ensemble of 32 E strategies each run for 5
s. These are our \emph{baseline} strategies.  Due to limited computational resources,
we do most of the experiments with the top 5 strategies that (greedily) cover
most solutions (\emph{top 5 greedy cover}). These are strategies
number 2, 8, 9, 26 and 28, henceforth called \emph{A}, \emph{B},
\emph{C}, \emph{D}, \emph{E}. In 5 s (in parallel) they together solve
21122 problems. We also evaluate these five strategies in 10 seconds,
jointly solving 21670 problems. The 21122 proofs yield over 100000
unique proof clauses that can be used for watchlist-based guidance in
our experiments.  We also use smaller datasets randomly sampled from
the full set of $57897$ problems to be able to explore more methods.
All problems are run on the same hardware\footnote{Intel(R) Xeon(R)
  CPU E5-2698 v3 @ 2.30GHz with 256G RAM.} and with the same memory
limits.

Each E strategy is specified as a frequency-weighted combination of
parameterized \emph{clause evaluation functions} (CEF) combined with a
selection of inference rules. Below we show a simplified example
strategy specifying the term ordering \emph{KBO}, and combining (with
weights 2 and 4) two CEFs made up of weight functions
\emph{Clauseweight} and \emph{FIFOWeight} and priority functions
\emph{DeferSOS} and
\emph{PreferWatchlist}. %
\begin{small}
\begin{verbatim}
-tKBO -H(2*Clauseweight(DeferSoS,20,9999,4),4*FIFOWeight(PreferWatchlist))
\end{verbatim}
\end{small}

\subsection{Watchlist Selection Methods}
\label{sect:watchlist}

We have experimented with several methods for
creation of static and dynamic watchlists. Typically we use only the proofs found by a particular baseline
strategy to construct the watchlists used for testing the guided version of that strategy. Using all $100000+$ proof clauses as a watchlist slows E down to 6 given clauses per second.  This is comparable to the speed of Prover9 with similarly large watchlists, but there are indexing methods that could speed this up. We have run several
smaller tests, but do not include this method in the evaluation
due to limited computational resources.
Instead, we select a smaller set of clauses. The methods are as follows:
\begin{enumerate}
\item[{\bf (art)}] Use all proof clauses from theorems in the problem's Mizar article\footnote{Excluding the current theorem.}. Such watchlist sizes range from $0$ to $4000$, which does not cause any significant slowdown of E.
\item[{\bf (freq)}] Use high-frequency proof clauses for static watchlists, i.e., clauses that appear in many proofs.
\item[{\bf (kNN-st)}] Use $k$-nearest neighbor ($k$-NN) learning to suggest useful
  static watchlists for each problem, based on symbol and term-based
  features~\cite{ckjujv-ijcai15} of the conjecture. This is very
  similar to the standard use of $k$-NN and other learners for premise
  selection. In more detail, we use symbols, walks of length 2 on
  formula trees and common subterms (with variables and skolem symbols
  unified). Each proof is turned into a multi-label training
  example, where the labels are the (serially numbered) clauses used
  in the proof, and the features are extracted from the conjecture.
\item[{\bf (kNN-dyn)}] Use $k$-NN in a similar way to suggest the most related proofs for
  dynamic watchlists. This is done in two iterations. 
  \begin{itemize}
  \item[{\bf (i)}] In the first
  iteration, only the conjecture-based similarity is used to select
  related problems and their proofs.
\item[{\bf (ii)}] 
 The second iteration then uses
  data mined from the proofs obtained with dynamic guidance in the
  first iteration. From each such proof $P$ we create a training
  example associating $P$'s conjecture features with the names of the
  proofs that matched (i.e., guided the inference of) the clauses
  needed in $P$. On this dataset we again train a $k$-NN learner, which recommends the most useful related proofs for guiding a particular conjecture.
  \end{itemize}
\end{enumerate}

\subsection{Using Watchlists in E Strategies}
\label{sect:use}

As described in Section~\ref{WatchE}, watchlist subsumption defines
the \texttt{PreferWatchlist} priority function that prioritizes clauses that
subsume at least one watchlist clause.  Below we
describe several ways to use this priority function and the newly
defined dynamic \texttt{PreferWatchlistRelevant} priority function and
its relevance-inheriting modifications. Each of them can additionally take the ``no-remove'' option, to keep subsumed watchlist clauses in the watchlist, allowing repeated matching by different clauses.
Preliminary testing has shown that just adding a single watchlist-based clause evaluation function (\emph{CEF}) to the baseline CEFs\footnote{Specifically we tried adding Defaultweight(PreferWatchlist) and ConjectureRelativeSymbolWeight(PreferWatchlist) with frequencies $1,2,5,10,20$ times that of the rest of the CEFs in the strategy.} %
is not as good as the methods defined below. In the rest of the paper we provide short names for the methods, such as \emph{prefA} (baseline strategy \emph{A} modified by the \emph{pref} method described below). %
\begin{enumerate}
\item \emph{evo}: the default heuristic strategy (Section~\ref{WatchE})
  evolved (genetically~\cite{SchaferS15}) for static watchlist use.
\item \emph{pref}: replace all priority functions in a baseline strategy with
  the \texttt{PreferWatch\-list} priority function.
  The resulting strategies look as follows:\\
  {\small\texttt{-H(2*Clauseweight(PreferWatchlist,20,9999,4),\\
      \phantom{-H(}4*FIFOWeight(PreferWatchlist))}}
\item \emph{const}: replace all priority functions in a baseline strategy with \texttt{ConstPrio},
  which assigns the same priority to all clauses, so all ranking is done by weight functions alone.
\item \emph{uwl}: always prefer clauses that match the watchlist,
  but use the baseline strategy's priority function otherwise\footnote{\emph{uwl} is implemented in E's source code as an option.}.
\item \emph{ska}: modify watchlist subsumption in E to treat all skolem symbols of the same arity as equal, thus widening the watchlist guidance. This can be used  with any strategy. In this paper it is used with \emph{pref}.
\item \emph{dyn}: replace all priority functions in a baseline strategy with \texttt{PreferWatchlist\-Relevant},
  which dynamically weights watchlist clauses (Section~\ref{Dynamic1}).
\item \emph{dyndec}: add the relevance inheritance mechanisms to \emph{dyn} (Section~\ref{Dynamic2}).
\end{enumerate}

\subsection{Evaluation}
\label{sect:performance}

First we measure the slowdown caused by larger static watchlists on the best baseline strategy
and a random sample of 10000 problems.
The results are shown in Table \ref{tab3}. %
We see that the speed significantly degrades with watchlists of size
10000, while 500-big watchlists incur only a small performance penalty.
\begin{table}[htbp]
\begin{footnotesize}
  \setlength\tabcolsep{4pt}
      \centering
      \begin{tabular}{c|rrrrrr}
 Size   &   10 &  100 &  256 &  512 & 1000 & 10000 \\ \hline
 proved & 3275 & 3275 & 3287 & 3283 & 3248 &  2912 \\
 PPS       & 8935 & 9528 & 8661 & 7288 & 4807 &   575 \\
      \end{tabular}
  \caption{\label{tab3}\small{Tests of the watchlist size influence (ordered by frequency) on a random sample of $10000$ problems using the "no-remove" option and one static watchlist with strategy \emph{prefA}. PPS is average processed clauses per second, a measure of E's speed.}}
\end{footnotesize}
\end{table}

%
%

Table~\ref{tab4} shows the 10 s evaluation of several static and dynamic methods on a random sample of $5000$ problems using article-based watchlists (method {\bf art} in Section \ref{sect:watchlist}).
For comparison, E's \emph{auto} strategy proves $1350$ of the problems in 10 s and its \emph{auto-schedule} proves $1629$.
Given 50 seconds the \emph{auto-schedule} proves $1744$ problems compared to our top 5 cover's 1964.

The first surprising result is that \emph{const} significantly outperforms the \emph{baseline}. This indicates that the old-style simple E priority functions may do more harm than good if they are allowed to override the more recent and sophisticated weight functions.
The \emph{ska} strategy performs best here and a variety of strategies provide better coverage. It's interesting to note that \emph{ska} and \emph{pref} overlap only on $1893$ problems. The original \emph{evo} strategy performs well, but lacks diversity.

\begin{table}[htp]
  \setlength\tabcolsep{4pt}
      \centering
  		\begin{tabular}{c|rcrrrrr}
      Strategy & baseline & const & pref & ska & dyn & evo & uwl\\ \hline
      A       & 1238 & 1493 & 1503 & \textbf{1510} & 1500 & 1303 & 1247 \\
      B       & 1255 & 1296 & 1315 & 1330 & 1316 & 1300 & 1277 \\
      C       & 1075 & 1166 & \textbf{1205} & 1183 & 1201 & 1068 & 1097 \\
      D       & 1102 & 1133 & 1176 & \textbf{1190} & 1175 & \textbf{1330} & 1132 \\
      E       & 1138 & \textbf{1141} & 1141 & 1153 & 1139 & 1070 & 1139 \\
      total   & 1853 & 1910 & 1931 & 1933 & 1922 & 1659 & 1868 \\
  		\end{tabular}
  	\caption{\label{tab4}\small{Article-based watchlist benchmark. A top 5 greedy cover proves 1964 problems (in bold).}}
\end{table}
Table~\ref{tab5a} briefly evaluates $k$-NN selection of watchlist clauses (method {\bf kNN-st} in Section \ref{sect:watchlist}) on a single strategy \emph{prefA}.
\begin{table}
      \centering
      \begin{tabular}{l|ccccc}
 Watchlist size   &   16 &   64 &  256 & 1024 & 2048 \\
 Proved & 1518 & 1531 & 1528 & 1532 & 1520 \\
      \end{tabular}
  \caption{\label{tab5a}\small{Evaluation of {\bf kNN-st} on prefA}}
\end{table}
Next we use k-NN to suggest watchlist proofs\footnote{All clauses in suggested proofs are used.} 
 (method {\bf kNN-dyn.i}) for \emph{pref} and \emph{dyn}.
Table~\ref{tab5} evaluates the influence of the number of related proofs loaded for the dynamic strategies. Interestingly, \emph{pref} outperforms \emph{dyn} almost everywhere but \emph{dyn}'s ensemble of strategies A-E generally performs best and the top 5 cover is better. We conclude that \emph{dyn}'s dynamic relevance weighting allows the strategies to diversify more.

Table~\ref{tab6} evaluates the top 5 greedy cover from Table \ref{tab5} on the full Mizar dataset, already showing significant improvement over the 21670 proofs produced by the 5 baseline strategies.
Based on proof data from a full-run of the top-5 greedy cover in Table \ref{tab6}, new k-NN proof suggestions were made (method {\bf kNN-dyn.ii}) and \emph{dyn}'s grid search re-run, see Table~\ref{tab7} and Table~\ref{tab8} for k-NN round 2 results.

We also test the relevance inheriting dynamic watchlist feature (\emph{dyndec}), primarily to determine if different proofs can be found. The results are shown in Table \ref{tab9}. This version adds $8$ problems to the top 5 greedy cover of all the strategies run on the $5000$ problem dataset, making it useful in a schedule despite lower performance alone. Table~\ref{tab10} shows this greedy cover, and then its evaluation on the full dataset. The 23192 problems proved by our new greedy cover is a 7\% improvement over the top 5 baseline strategies.

\begin{table}[htbp]
  \setlength\tabcolsep{4pt}
      \centering
      \begin{tabular}{c|rrrrr|r}
        size & dynA & dynB & dynC & dynD & dynE & total\\ \hline
        4    & 1531 & 1352 & 1235 & 1194 & \textbf{1165} & 1957\\
        8    & 1543 & 1366 & \textbf{1253} & 1188 & 1170 & 1956\\
        16   & 1529 & 1357 & 1224 & \textbf{1218} & 1185 & 1951\\
        32   & \textbf{1546} & 1373 & 1240 & 1218 & 1188 & 1962\\
        64   & 1535 & \textbf{1376} & 1216 & 1215 & 1166 & 1935\\
        128  & 1506 & 1351 & 1195 & 1214 & 1147 & 1907\\
        1024 & 1108 & 963  & 710  & 943  & 765 & 1404\\
       \\ \hline
        size & prefA & prefB & prefC & prefD & prefE & total\\ \hline
        4    & 1539 & 1369 & 1210 & 1220 & \textbf{1159} & 1944\\
        8    & 1554 & 1385 & \textbf{1219} & 1240 & 1168 & 1941\\
        16   & \textbf{1572} & 1405 & 1225 & 1254 & 1180 & 1952\\
        32   & 1568 & 1412 & 1231 & \textbf{1271} & 1190 & 1958\\
        64   & 1567 & 1402 & 1228 & 1262 & 1172 & 1952\\
        128  & \textbf{1552} & 1388 & 1210 & 1248 & 1160 & 1934\\
        1024 & 1195 & 1061 & 791  & 991  & 806  & 1501\\
      \end{tabular}
  \caption{\label{tab5}\small{k-NN proof recommendation watchlists ({\bf kNN-dyn.i}) for \emph{dyn} \emph{pref}. Size is number of proofs, averaging 40 clauses per proof.  A top 5 greedy cover of \emph{dyn} proves 1972 and \emph{pref} proves 1959 (in bold).}}
\end{table}

\begin{table}[htp]
  \setlength\tabcolsep{4pt}
      \centering
      \begin{tabular}{c|rrrrr}
         & dynA\_32 & dynC\_8 & dynD\_16 & dynE\_4 & dynB\_64\\ \hline
        added & 17964 & 2531        & 1024        & 760        & 282 \\
        total & 17964 & 14014       & 14294       & 13449      & 16175\\

      \end{tabular}
  \caption{\label{tab6}\small{K-NN round 1 greedy cover on full dataset and proofs added by each successive strategy for a total of $22579$. dynA\_32 means strategy \emph{dynA} using 32 proof watchlists.}}
\end{table}
\begin{table}[htbp]
  \setlength\tabcolsep{4pt}
      \centering
      \begin{tabular}{c|rrrrr|cc}
        size & dyn2A & dyn2B & dyn2C & dyn2D & dyn2E & total & round 1 total\\ \hline
        4    & 1539    & \textbf{1368} & 1235    & 1209    & \textbf{1179} & 1961 & 1957\\
        8    & 1554    & 1376    & 1253    & 1217    & 1183    & 1971 & 1956\\
        16   & \textbf{1565} & 1382    & \textbf{1256} & 1221    & 1181    & 1972 & 1951\\
        32   & 1557    & 1383    & 1252    & \textbf{1227}    & 1182    & 1968 & 1962\\
        64   & 1545    & 1385    & 1244    & 1222    & 1171    & 1963 & 1935\\
        128  & 1531    & 1374    & 1221    & 1227    & 1171    & 1941 & 1907\\
      \end{tabular}
  \caption{\label{tab7}\small{Problems proved by round 2 k-NN proof suggestions ({\bf kNN-dyn.ii}). The top 5 greedy cover proves $1981$ problems (in bold). \emph{dyn2A} means \emph{dynA} run on the 2nd iteration of k-NN suggestions.}}
\end{table}
\begin{table}[htp]
  \setlength\tabcolsep{4pt}
      \centering
      \begin{tabular}{c|rrrrr}
         & dyn2A\_16 & dyn2C\_16 & dyn2D\_32 & dyn2E\_4 & dyn2B\_4\\ \hline
        total & 18583   & 14486        & 14720       & 13532        & 16244\\
        added & 18583   & 2553         & 1007        & 599          & 254 \\
      \end{tabular}
  \caption{\label{tab8}\small{K-NN round 2 greedy cover on full dataset and proofs added by each successive strategy for a total of $22996$}}
\end{table}
\begin{table}[htp]
  \setlength\tabcolsep{4pt}
      \centering
      \begin{tabular}{c|rrrrr|c}
        size & dyndec2A & dyndec2B & dyndec2C & dyndec2D & dyndec2E & total\\ \hline
        4    & \textbf{1432} & 1354 & \textbf{1184} & \textbf{1203} & \textbf{1152} & 1885 \\
        16   & 1384 & 1316 & 1176 & 1221 & 1140 & 1846 \\
        32   & 1381 & 1309 & 1157 & 1209 & 1133 & 1820 \\
        128  & 1326 & \textbf{1295} & 1127 & 1172 & 1082 & 1769 \\
      \end{tabular}
  \caption{\label{tab9}\small{Problems proved by round 2 k-NN proof suggestions with \emph{dyndec}. The top 5 greedy cover proves $1898$ problems (in bold).}}
\end{table}
\begin{table}[htp]
  \setlength\tabcolsep{4pt}
      \centering
      \begin{tabular}{c|ccccc}
        total & dyn2A\_16 & dyn2C\_16 & dyndec2D\_16 & dyn2E\_4 & dyndec2A\_128\\ \hline
        2007  & 1565   & 230         & 97        & 68          & 47 \\ \hline \hline
        23192 & 18583  & 2553        & 1050      & 584         & 422 \\
        23192 & 18583  & 14486       & 14514     & 13532       & 15916 \\
      \end{tabular}
  \caption{\label{tab10}\small{Top: Cumulative sum of the $5000$ test set greedy cover. The k-NN based dynamic watchlist methods dominate, improving by $2.1\%$ over the baseline and article-based watchlist strategy greedy cover of $1964$ (Table~\ref{tab4}). Bottom: Greedy cover run on the full dataset, cumulative and total proved.}}
\end{table}

%

%
%

%

%
%
%
%
%
%
%
%
%
%
%
%
%
%
%
%
%
%
%
%
%
%
%
%
%
%
%
%
%
%
%
%
%
%
%
%
%
%
%

%

\section{Examples}
\label{Examples}
The Mizar theorem \texttt{YELLOW\_5:36}\footnote{\url{http://grid01.ciirc.cvut.cz/~mptp/7.13.01_4.181.1147/html/yellow_5\#T36}} states De Morgan's laws for Boolean lattices:
\begin{lstlisting}[language=Mizar,basicstyle=\ttfamily\scriptsize]
theorem Th36: :: YELLOW_5:36
for L being non empty Boolean RelStr for a, b being Element of L
holds ( 'not' (a "\/" b) = ('not' a) "/\" ('not' b)
         & 'not' (a "/\" b) = ('not' a) "\/" ('not' b) )
\end{lstlisting}
Using 32 related proofs results in 2220 clauses placed on the watchlists.
The dynamically guided proof search takes 5218 (nontrivial) given clause loops done in 2 s and the resulting ATP proof is 436 inferences long.
There are 194 given clauses that match the watchlist during the proof search and 120 (61.8\%)
of them end up being part of the proof. I.e., 27.5\% of the proof
consists of steps guided by the watchlist mechanism. The proof search using the same settings, but without the watchlist
takes 6550 nontrivial given clause loops (25.5\% more).
The proof of the theorem \texttt{WAYBEL\_1:85}\footnote{\url{http://grid01.ciirc.cvut.cz/~mptp/7.13.01_4.181.1147/html/waybel_1\#T85}} is considerably used for this guidance:
\begin{lstlisting}[language=Mizar,basicstyle=\ttfamily\scriptsize]
theorem :: WAYBEL_1:85
for H being non empty lower-bounded RelStr st H is Heyting holds
for a, b being Element of H holds 'not' (a "/\" b) >= ('not' a) "\/" ('not' b)
\end{lstlisting}
Note that this proof is done under the weaker assumptions of H being
lower bounded and Heyting, rather than being Boolean. Yet, 62 (80.5\%) of
the 77 clauses from the proof of \texttt{WAYBEL\_1:85} are eventually matched
during the proof search. 38 (49.4\%) of these 77 clauses are used in the proof of \texttt{YELLOW\_5:36}.
In Table~\ref{tabprog} we show the final state of proof progress for the 32 loaded proofs after the last non empty clause matched the watchlist. For each we show both the computed ratio and the number of matched and all clauses.
\begin{table}
  \setlength\tabcolsep{5.5pt}
\begin{tabular}{ccc|ccc|ccc|ccc}
 0& 0.438 &      42/96& 1& 0.727 &      56/77& 2& 0.865 &      45/52& 3& 0.360 &       9/25\\
 4& 0.750 &      51/68& 5& 0.259 &       7/27& 6& 0.805 &      62/77& 7& 0.302 &      73/242\\
 8& 0.652 &      15/23& 9& 0.286 &       8/28& 10& 0.259 &       7/27& 11& 0.338 &      24/71\\
12& 0.680 &      17/25& 13& 0.509 &      27/53& 14& 0.357 &      10/28& 15& 0.568 &      25/44\\
16& 0.703 &      52/74& 17& 0.029 &       8/272& 18& 0.379 &      33/87& 19& 0.424 &      14/33\\
20& 0.471 &      16/34& 21& 0.323 &      20/62& 22& 0.333 &       7/21& 23& 0.520 &      26/50\\
24& 0.524 &      22/42& 25& 0.523 &      45/86& 26& 0.462 &       6/13& 27& 0.370 &      20/54\\
28& 0.411 &      30/73& 29& 0.364 &      20/55& 30& 0.571 &      16/28& 31& 0.357 &      10/28\\
\end{tabular}
\caption{\label{tabprog}{Final state of the proof progress for the (serially numbered) 32 proofs loaded to guide the proof of \texttt{YELLOW\_5:36}. We show the
computed ratio and the number of matched and all clauses.}}
\end{table}

An example of a theorem that can be proved in 1.2 s with guidance but cannot be proved in 10 s with any unguided method is
the following theorem \texttt{BOOLEALG:62}\footnote{\url{http://grid01.ciirc.cvut.cz/~mptp/7.13.01_4.181.1147/html/boolealg\#T62}} about the symmetric difference in Boolean lattices:
\begin{lstlisting}[language=Mizar,basicstyle=\ttfamily\scriptsize]
for L being B_Lattice
for X, Y being Element of L holds (X \+\ Y) \+\ (X "/\" Y) = X "\/" Y
\end{lstlisting}
Using 32 related proofs results in 2768 clauses placed on the watchlists.
The proof search then takes  4748 (nontrivial) given clause loops and the watchlist-guided ATP proof is 633 inferences long.
There are 613 given clauses that match the watchlist during the proof search and 266 (43.4\%)
of them end up being part of the proof. I.e., 42\% of the proof
consists of steps guided by the watchlist mechanism. Among the theorems whose proofs are most useful for the guidance are the following theorems \texttt{LATTICES:23}\footnote{\url{http://grid01.ciirc.cvut.cz/~mptp/7.13.01_4.181.1147/html/lattices\#T23}}, \texttt{BOOLEALG:33}\footnote{\url{http://grid01.ciirc.cvut.cz/~mptp/7.13.01_4.181.1147/html/boolealg\#T33}} and \texttt{BOOLEALG:54}\footnote{\url{http://grid01.ciirc.cvut.cz/~mptp/7.13.01_4.181.1147/html/boolealg\#T54}} on Boolean lattices:
\begin{lstlisting}[language=Mizar,basicstyle=\ttfamily\scriptsize]
theorem Th23: :: LATTICES:23
for L being B_Lattice
for a, b being Element of L holds (a "/\" b)` = a` "\/" b`

theorem Th33: :: BOOLEALG:33
for L being B_Lattice for X, Y being Element of L holds X \ (X "/\" Y) = X \ Y

theorem :: BOOLEALG:54
for L being B_Lattice for X, Y being Element of L
st X` "\/" Y` = X "\/" Y & X misses X` & Y misses Y`
holds  X = Y` & Y = X`
\end{lstlisting}

Finally, we show several theorems\begin{footnoterange}
\footnote{\url{http://grid01.ciirc.cvut.cz/~mptp/7.13.01_4.181.1147/html/boolealg\#T68}}\footnote{\url{http://grid01.ciirc.cvut.cz/~mptp/7.13.01_4.181.1147/html/closure1\#T21}}\footnote{\url{http://grid01.ciirc.cvut.cz/~mptp/7.13.01_4.181.1147/html/bcialg_4\#T44}}\footnote{\url{http://grid01.ciirc.cvut.cz/~mptp/7.13.01_4.181.1147/html/xxreal_3\#T67}}
\end{footnoterange}
with nontrivial Mizar proofs and relatively long ATP proofs obtained with significant guidance. These theorems cannot be proved by any other method used in this work.
\begin{lstlisting}[language=Mizar,basicstyle=\ttfamily\scriptsize]
theorem :: BOOLEALG:68
for L being B_Lattice for X, Y being Element of L
holds (X \+\ Y)` = (X "/\" Y) "\/" ((X`) "/\" (Y`))

theorem :: CLOSURE1:21
for I being set for M being ManySortedSet of I
for P, R being MSSetOp of M st P is monotonic & R is monotonic
holds P ** R is monotonic

theorem :: BCIALG_4:44
for X being commutative BCK-Algebra_with_Condition(S)
for a, b, c being Element of X st Condition_S (a,b) c= Initial_section c holds
for x being Element of Condition_S (a,b) holds x <= c \ ((c \ a) \ b)

theorem :: XXREAL_3:67
for f, g being ext-real number holds (f * g)"=(f") * (g")
\end{lstlisting}

\section{Related Work and Possible Extensions}
\label{Related}
The closest related work %
is the hint
guidance in Otter and Prover9. Our focus is however on large ITP-style
theories with large signatures and heterogeneous facts and proofs spanning various areas of mathematics. This
motivates using machine learning for reducing the size of the static
watchlists and the implementation of the dynamic watchlist mechanisms.
Several implementations of internal proof search guidance using
statistical learning have been mentioned in Sections~\ref{Intro} and
\ref{State}. In both the tableau-based systems and the tactical ITP
systems the statistical learning guidance benefits from a compact and directly
usable notion of proof state, which is not immediately available in
saturation-style ATP.

By delegating the notion of similarity to subsumption %
we are relying on fast, crisp and well-known symbolic ATP
mechanisms. This has advantages as well as disadvantages. Compared
to the ENIGMA~\cite{JakubuvU17a} and neural~\cite{LoosISK17} statistical guiding methods,
the subsumption-based notion of clause similarity is not feature-based or learned.
This similarity relation is crisp and sparser compared to the similarity
relations induced by the statistical methods. The proof guidance is limited when
no derived clauses subsume any of
the loaded proof clauses. This can be countered by loading a high
number of proofs and widening (or softening) the similarity
relation in various approximate ways. On the other hand, subsumption is fast
compared to the deep neural methods (see~\cite{LoosISK17}) and enjoys
clear guarantees of the underlying symbolic calculus. For example, when
all the (non empty) clauses from a loaded related proof have been subsumed in
the current proof search, it is clear that the current proof search is
successfully finished.

A clear novelty is the focusing of the proof search towards the
(possibly implausible) inferences needed for completing the loaded
proofs.  Existing statistical guiding methods will fail to notice such
opportunities, and the static watchlist guidance has no way of distinguishing
the watchlist matchers that lead faster to proof completion.
In a way this mechanism resembles the feedback obtained by Monte Carlo exploration,
where a seemingly statistically unlikely decision can be made, based
on many rollouts and averaging of their results.
Instead, we rely here on a database of previous proofs, similar
to previously played and finished games. The newly introduced heuristic proof
search (proof progress) representation may however enable
further experiments with Monte Carlo guidance.

%
%
%
%

%
%
%
%

%

%
%

%
%
%
%
%

%
%
%
%
%

%

%

\subsection{Possible Extensions}
Several extensions have been already discussed above. We list the most obvious.
\paragraph{\textbf{More sophisticated progress metrics}:} The current proof-progress
criterion may be too crude. Subsuming all the \emph{initial} clauses of a
related proof is unlikely until the empty clause is derived. In
general, a large part of a related proof may not be needed once the
right clauses in the ``middle of the proof'' are subsumed by the
current proof search. A better
proof-progress metric would
compute the smallest number of proof
clauses that are still needed to entail the contradiction.
This is
achievable, however more technically involved, also due to issues such as
rewriting of the watchlist clauses during the current proof search.

\paragraph{\textbf{Clause re-evaluation based on the evolving proof relevance}:}
As more and more watchlist clauses are matched, the proof relevance of
the clauses generated earlier should be updated to mirror the current
state. This is in general expensive, so it could be done after each
$N$ given clause loops or after a significant number of watchlist
matchings. An alternative is to add corresponding indexing mechanisms
to the set of generated clauses, which will immediately reorder them
in the evaluation queues based on the proof relevance updates.

\paragraph{\textbf{More abstract/approximate matching}:} Instead of
the strict notion of subsumption, more abstract or heuristic matching
methods could be used. An interesting symbolic method to consider is
matching modulo symbol alignments~\cite{GauthierK14}. A number of
approximate methods are already used by the above mentioned statistical guiding methods.

\paragraph{\textbf{Adding statistical methods for clause guidance}:} Instead of
using only hard-coded watchlist-style heuristics for focusing inferences, a statistical (e.g. ENIGMA-style)
clause evaluation model could be trained
by adding the vector of proof completion ratios to the currently used ENIGMA features.

\section{Conclusion}
\label{sect:conclusion}
The portfolio of new proof guiding methods developed here
significantly improves E's standard portfolio of strategies, and also
the previous best set of strategies invented for Mizar by evolutionary
methods. The best combination of five new strategies run in
parallel for 10 seconds (a reasonable hammering time) will prove over
7\% more Mizar problems than the previous best combination of five
non-watchlist strategies. Improvement over E's standard portfolio
is much higher.
Even though we focus on developing the strongest portfolio rather than
a single best method, it is clear that the best guided versions also
significantly improve over their non-guided counterparts. This
improvement for the best new strategy (\texttt{dyn2A} used with 16
most relevant proofs) is 26.5\% ($= 18583 / 14693$). These are
relatively high improvements in automated theorem proving.

We have shown that the new dynamic methods based on the idea of proof
completion ratios improve over the static watchlist guidance.  We have
also shown that as usual with learning-based guidance, iterating the
methods to produce more proofs leads to stronger methods in the next
iteration. The first experiments with widening the watchlist-based
guidance by relatively simple inheritance mechanisms seem quite
promising, contributing many new proofs. A number of extensions and
experiments with guiding saturation-style proving have been opened for
future research. We believe that various extensions of the compact and
evolving heuristic representation of saturation-style proof search as
introduced here will turn out to be of great importance for further
development of learning-based saturation provers.

\section{Acknowledgments}
We thank Bob Veroff for many enlightening explanations and discussions
of the watchlist mechanisms in Otter and Prover9. His
``industry-grade'' projects that prove open and interesting
mathematical conjectures with hints and proof sketches have been a great
sort of inspiration for this work.
\bibliographystyle{abbrv}
\bibliography{ate11.bib,stsbib.bib}

\end{document}